\documentclass[12pt]{article}

\usepackage{amsmath}
\usepackage{amssymb}
\usepackage{latexsym}
\usepackage{epsfig}
\usepackage{epsf}
\usepackage{fullpage}
\usepackage{setspace}
\usepackage{multirow}
\usepackage{multicol}
\usepackage{subfigure} 
\usepackage{dcolumn}
\usepackage{endfloat}
\usepackage{longtable}
\newcolumntype{.}{D{.}{.}{4}}
\DeclareMathOperator*{\prodkK}{\prod_{k=1}^K}

\DeclareMathOperator*{\tauk}{\tau_k^2}

\DeclareMathOperator*{\lamo}{\lambda_1^2}
\DeclareMathOperator*{\lamt}{\lambda_2^2}

\DeclareMathOperator*{\argmin}{arg\;min}  
\DeclareMathOperator*{\argmax}{arg\;max}  

\DeclareMathOperator*{\mlikelihood}{\sum_{\ell = 1}^n || \mathbf{W}^T\mathbf{x}_{\ell} - \mathbf{y}_{\ell}||_2^2}

\DeclareMathOperator*{\W}{\mathbf{W}}
\DeclareMathOperator*{\Y}{\mathbf{Y}}

\DeclareMathOperator*{\yl}{\mathbf{y}_{\ell}}

\DeclareMathOperator*{\omei}{\omega_i^2}

\DeclareMathOperator*{\sig}{\sigma^2}

\DeclareMathOperator*{\vtau}{\underset{\sim}{\mathbf{\tau}^2}}

\DeclareMathOperator*{\prodid}{\prod_{i=1}^d}

\def \vW{\mbox{\boldmath $W$ \unboldmath}\!\!}
\def \vB{\mbox{\boldmath $B$ \unboldmath}\!\!}

\def \vSigma{\mbox{\boldmath $\Sigma$ \unboldmath}\!\!}

\def \vomega{\mbox{\boldmath $\omega$ \unboldmath}\!\!}

\def \vA{\mbox{\boldmath $A$ \unboldmath}\!\!}

\def \vtau{\mbox{\boldmath $\tau$ \unboldmath}\!\!}
\def \vomega{\mbox{\boldmath $\omega$ \unboldmath}\!\!}

\def \vY{\mbox{\boldmath $Y$ \unboldmath}\!\!}
\def \vy{\mbox{\boldmath $y$ \unboldmath}\!\!}

\def \vx{\mbox{\boldmath $x$ \unboldmath}\!\!}
\def \vX{\mbox{\boldmath $X$ \unboldmath}\!\!}

\begin{document}

\title{Regularization Parameter Selection for a Bayesian Multi-Level Group Lasso Regression Model with Application to Imaging Genomics}


\author{Farouk S. Nathoo$^{*}$, Keelin Greenlaw, Mary Lesperance\\
Department of Mathematics and Statistics,
University of Victoria\\
Victoria, Canada\\
$^{*}$nathoo@uvic.ca\\}
\date{}
\maketitle

\begin{abstract}
We investigate the choice of tuning parameters for a Bayesian multi-level group lasso model developed for the joint analysis of neuroimaging and genetic data. The regression model we consider relates multivariate phenotypes consisting of brain summary measures (volumetric and cortical thickness values) to single nucleotide polymorphism (SNPs) data and imposes penalization at two nested levels, the first corresponding to genes and the second corresponding to SNPs. Associated with each level in the penalty is a tuning parameter which corresponds to a hyperparameter in the hierarchical Bayesian formulation. Following previous work on Bayesian lassos we consider the estimation of tuning parameters through either hierarchical Bayes based on hyperpriors and Gibbs sampling or through empirical Bayes based on maximizing the marginal likelihood using a Monte Carlo EM algorithm. For the specific model under consideration we find that these approaches can lead to severe overshrinkage of the regression parameter estimates in the high-dimensional setting or when the genetic effects are weak. We demonstrate these problems through simulation examples and study an approximation to the marginal likelihood which sheds light on the cause of this problem. We then suggest an alternative approach based on the widely applicable information criterion (WAIC), an asymptotic approximation to leave-one-out cross-validation that can be computed conveniently  within an MCMC framework.  
\end{abstract}

\section{Introduction}

Imaging genomics involves the joint analysis of neuroimaging and genetic data with the goal of examining genetic risk variants that may relate to the structure or function of the brain. We focus here specifically on multivariate regression analysis where the response vector comprises brain imaging phenotypes that we wish to relate to high-throughput single nucleotide polymorphism (SNP) data. We have developed a Bayesian approach for regression analysis in this setting based on a continuous shrinkage prior for the regression coefficients that induces dependence in the coefficients corresponding to SNPs within the same gene, and across different components of the imaging phenotypes. The specific purpose of this contribution is to discuss methods for choosing the tuning parameters of this model.

Let $\yl = (y_{\ell 1}, \dots , y_{ \ell c})' \hspace{5pt}$ denote the imaging phenotype summarizing the structure of the brain over $c$ regions of interest (ROIs) for subject $\ell$, $\ell = 1,\dots, n$. The corresponding genetic data are denoted by $\mathbf{x}_{\ell} = (x_{\ell 1}, \dots , x_{\ell d})', \hspace{5pt}\ell = 1,\dots, n$, where these data comprise information on $d$ SNPs, and $x_{\ell j} \in \{0,1,2\}$ is the number of minor alleles for the $j^{th}$ SNP. We further assume that the set of SNPs can be partitioned into $K$ groups, for example $K$ genes, and we let  $\pi_k, k = 1,2, \dots, K$, denote the set containing the SNP indices corresponding to the $k^{th}$ gene, and $m_{k}$ denotes the cardinality of $\pi_k$. We assume that $E(\mathbf{y_{\ell}}) = \mathbf{W}^{T}\mathbf{x}_\ell, \hspace{5pt} \ell = 1,\dots, n$, where $\mathbf{W}$ is a $d$ x $c$ matrix, with each row characterizing the association between a given SNP and the brain summary measures across all ROIs. The specific model we consider is motivated by the group sparse multi-task and feature selection estimator proposed in \cite{wang2012identifying}
\begin{small}
\begin{equation}
\label{Wang estimator} 
\hat{\mathbf{W}} = \underset{\mathbf{W}}\argmin \left\lbrace  \mlikelihood  +\gamma_1 ||\mathbf{W}||_{G_{2,1}} +\gamma_2 ||\mathbf{W}||_{l_{2,1}}   \right\rbrace 
\end{equation}
\end{small}
where $\gamma_1$ and $\gamma_2$ are regularization parameters weighting a $G_{2,1}$-norm penalty $||\mathbf{W}||_{G_{2,1}} = \sum_{k=1}^K  \sqrt{\sum_{i \in \pi_k}  \sum_{j=1}^c w_{ij}^2 }$ and an $\ell_{2,1}$-norm penalty $||\mathbf{W}||_{l_{2,1}} = \sum_{i=1}^{d}  \sqrt{ \sum_{j=1}^c w_{ij}^2}$ respectively.

The penalty encourages group sparsity at two nested levels, at the level of genes through the $G_{2,1}$-norm penalty and at the level of SNPs through the $\ell_{2,1}$-norm penalty. We have developed an extension of this approach based on a Bayesian formulation where the posterior mode conditional on hyperparameters is equal to (\ref{Wang estimator}). An advantage of considering a Bayesian approach in this context is that the posterior distribution furnishes not only a point estimate but also a mechanism for conducting statistical inference, which includes the construction of interval estimates. It thus allows for uncertainty quantification over and above point estimation (see e.g. \cite{park2008bayesian}, \cite{kyung2010penalized}). The Bayesian model can be specified through a three-level hierarchy
$$\yl |\mathbf{W},\sigma^2  \stackrel{ind}{\sim} MVN_c (\vW^{T} \vx_\ell \: , \: \sigma^2I_c) \hspace{8pt} \ell=1, \dots, n;$$
$$w_{ij}\; |\; \sigma^2, \; \vtau^{2}, \; \vomega^{2}\; \stackrel{ind}{\sim} \; N \left( 0, \;\sigma^2 (  \frac{1}{\tau_{k(i)}^2}\; +\; \frac{1}{\omega_i^2} )^{-1}\: \right),$$
where $k(i) \in \{1,\dots,K\}$ denotes the gene associated with the $i^{th}$ SNP and the model involves continuous scale mixing variables $\vtau^{2} = (\tau_{1}^{2},\dots,\tau_{K}^{2})'$ and $\vomega^{2} = (\omega_{1}^{2},\dots,\omega_{d}^{2})'$ distributed according to the density $p(\vtau^{2},\vomega^{2}|\lambda_{1}^{2},\lambda_{2}^{2})$
\begin{align}
\label{scale-mix2}
\begin{split}
\propto&\prodkK  \left(\frac{\lambda_1^2}{2}\right)^{\left(\frac{m_kc + 1}{2}\right)}(\tau_k^2)^{\left(\frac{m_kc + 1}{2}\right) -1 } \exp \left\lbrace -\left(\frac{\lambda_1^2}{2}\right) \tau_k^2 \right\rbrace \\
&\times  \prod_{i \in \pi_{k}}   \left(\frac{\lambda_2^2}{2}\right)^{\left(\frac{c + 1}{2}\right)}(\omega_i^{2})^{\left(\frac{c + 1}{2}\right) -1 } \exp \left\lbrace -\left(\frac{\lambda_2^2}{2}\right) \omega_{i}^{2} \right\rbrace\\
& \times (\tau_{k}^{2}+ \omega_{i}^{2})^{-\frac{c}{2}}.  
\end{split}
\end{align}
The form (\ref{scale-mix2}) is required to ensure that the posterior distribution obtained after marginalizing over $\vtau^{2}$ and $\vomega^{2}$, $[\vW|\vY, \sigma^{2}, \lambda_{1}^{2},\lambda_{2}^{2}]$, has mode exactly (\ref{Wang estimator}) with $\gamma_{1} = 2\sigma \lambda_{1}$, $\gamma_{2} = 2  \sigma \lambda_{2}$, and $\vY = (\vy_{1}^{T},\dots,\vy_{n}^{T})^{T}$. 

The parameter $\sigma^{2}$ is assigned a proper inverse-Gamma prior
$\sigma^2 \; \sim \; Inv-Gamma (a_\sigma, b_\sigma )$
where setting $a_\sigma=2, b_\sigma=1$ yields a fairly weakly-informative prior in many (but not all) settings. The hierarchical model has a conjugacy structure that facilitates posterior simulation using a Gibbs sampling algorithm. As the normalizing constant associated with (\ref{scale-mix2}) is not known and may not exist we work with the unnormalized form which yields proper full conditional distributions having standard form. Our focus of inference does not lie with the scale mixing variables themselves, rather, the use of the scale mixture representation is a computational device that leads to a fairly straightforward Gibbs sampling algorithm which enables us to draw from the marginal posterior distribution $[\vW|\vY,\lambda_{1}^{2},\lambda_{2}^{2}]$.  

\section{Tuning Parameter Selection}

The tuning parameters can be estimated by maximizing the marginal likelihood $\hat{\lambda}_{1}^{2}, \;   \hat{\lambda}_{2}^{2} \;= \; \underset{\lambda_{1}^{2}, \lambda_{2}^{2}} \argmax \;p\: ( \Y | \lambda_{1}^{2}, \lambda_{2}^{2})$
$
 = \; \underset{\lambda_{1}^{2}, \lambda_{2}^{2}} \argmax \;  \int_\Theta p\: ( \Y , \Theta \; | \lambda_{1}^{2}, \lambda_{2}^{2}) \; d \Theta
$
where $\Theta = ( \W , \vtau^{2}, \vomega^{2}, \sigma^{2})$. This optimization problem can be solved with a Monte Carlo Expectation Maximization (MCEM) algorithm \cite{park2008bayesian}.

The fully Bayes alternative for handling the tuning parameters is to assign these parameters hyper-priors and sample them from the corresponding posterior distribution $p(\lambda_{1}^{2},\lambda_{2}^{2}|\Y) \propto p( \Y | \lambda_{1}^{2}, \lambda_{2}^{2})p( \lambda_{1}^{2})p(\lambda_{2}^{2})$, where the posterior samples can be obtained by adding additional update steps to the Gibbs sampling algorithm. It is computationally convenient in this case to assign conditionally conjugate priors $\lambda_{i}^{2} \stackrel{ind}{\sim} \text{Gamma}(r_{i},\delta_{i}), i=1,2$.  

As a third alternative cross-validation (CV) can be employed. A problem with the use of CV when MCMC runs are required to fit the model is that an extremely large number of parallel runs are needed to cover all points on the grid for each possible split of the data. To avoid some of this computational burden we approximate leave-one-out CV using the WAIC \cite{watanabe2010asymptotic},
$$
WAIC = -2\sum_{l=1}^{n}\log E_{\W,\sigma^{2}}[p(\yl | \W,\sigma^{2}) | \mathbf{y}_{1}, \dots, \mathbf{y}_{n}] 
+ 2\sum_{l=1}^{n} VAR_{\W,\sigma^{2}}[\log p(\yl | \W,\sigma^{2}) | \mathbf{y}_{1}, \dots, \mathbf{y}_{n}]
$$
where $p(\yl | \W,\sigma^{2})$ is the Gaussian density function for $[\yl | \W,\sigma^{2}]$ and the computation is based on the output of the Gibbs sampler at each point on a grid of values for $\lambda_{1}^{2}$ and $\lambda_{2}^{2}$. The values of $\lambda_{1}^{2}$ and $\lambda_{2}^{2}$ are then chosen as those values that minimize the WAIC.

Conceptually we find the fully Bayesian approach to be the most appealing of the three possibilities mentioned above, as this is the only approach out of the three that takes account of the posterior uncertainty associated with $\lambda_{1}^{2}$ and $\lambda_{2}^{2}$. In practice, however, we have found that the fully Bayes and empirical Bayes approaches, both of which are primarily governed by the shape of marginal likelihood $p\: ( \Y | \lambda_{1}^{2}, \lambda_{2}^{2})$, can lead to problems \emph{for our proposed model} in settings where the number of SNPs is large relative to $n$, or when the effect sizes are weak. We have observed these problems to occur quite generally but they are illustrated here with two simple examples based on simulated data. In the first example we set the number of SNPs to be $d=200$, the number of genes to be $K=20$, the number of imaging phenotypes to be $c=5$, and the number of subjects to be $n=500$. In the second example we increase the number of SNPs to $d=1500$, the number of genes to $K=150$, and again set $c=5$, and $n=500$ and the SNP covariate data $\vx_{\ell}, \ell = 1,\dots,n$, are generated independently and uniformly from $\{0,1,2\}$. In each example we simulate a single realization from a slightly perturbed version the model where for ease of simulation it is assumed that 
$$
\tau_k^2 \; | \; \lambda_{1}^{2} \; \stackrel{ind}{\sim} \; Gamma \left( \frac{m_k\*c+1}{2} \;, \; \frac{\lambda_{1}^{2}}{2} \right),\,\,\, k=1,\dots,K, 
$$ and 
$$
\omega_i^2 \; | \; \lambda_{2}^{2} \; \stackrel{ind}{\sim} \; Gamma \left( \frac{c+1}{2} \;, \; \frac{\lambda_{2}^{2}}{2} \right), \,\,\, i=1,\dots,d, 
$$ and we set $\lambda_{1}^{2} = 2$, $\lambda_{2}^{2} = 2$, and $\sigma^{2} = 2$. We emphasize that this simplifying assumption is made only when simulating the data, but not when fitting the actual model to this data.

For the first case where $d=200$, Figure 1, panel (a), shows the output of the MCEM algorithm (the estimate after each iteration) based on four different sets of initial values for $\lambda_{1}^{2}$. In this case we see that the algorithm converges to the same solution for all four sets of initial values, and this is the empirical Bayes estimate. Similarly, the MCEM output for $\lambda_{2}^{2}$ (not shown) also indicates convergence. For the fully Bayes approach, Figure 1, panel (b), shows the output of the Gibbs sampler for $\lambda_{1}^{2}$ for five parallel sampling chains initialized at different states. The Gibbs sampler mixes well and we note that the stationary distribution is centred close to the point estimates obtained from the MCEM algorithm. The Gibbs sampling output for $\lambda_{2}^{2}$ (not shown) behaves in a similar manner.
\begin{figure}[htbp]
\centering
\begin{tabular}{cc}
\includegraphics[scale=0.28]{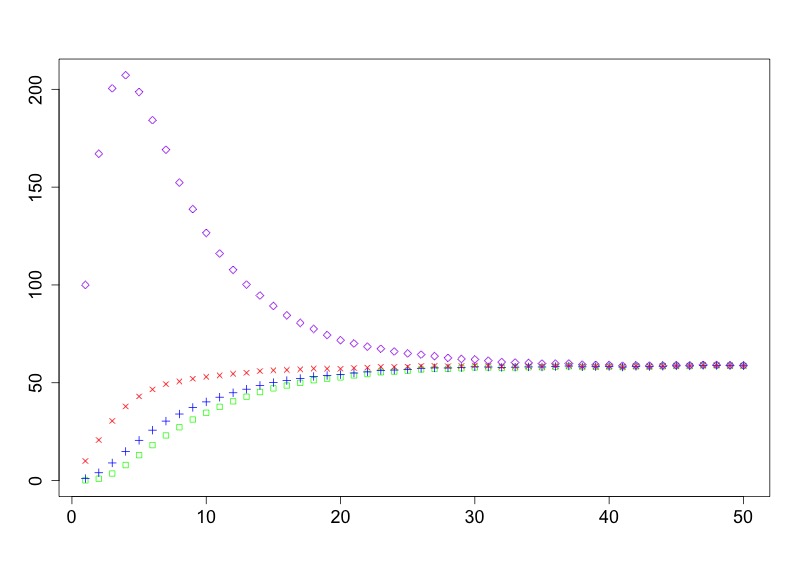}&
\includegraphics[scale=0.28]{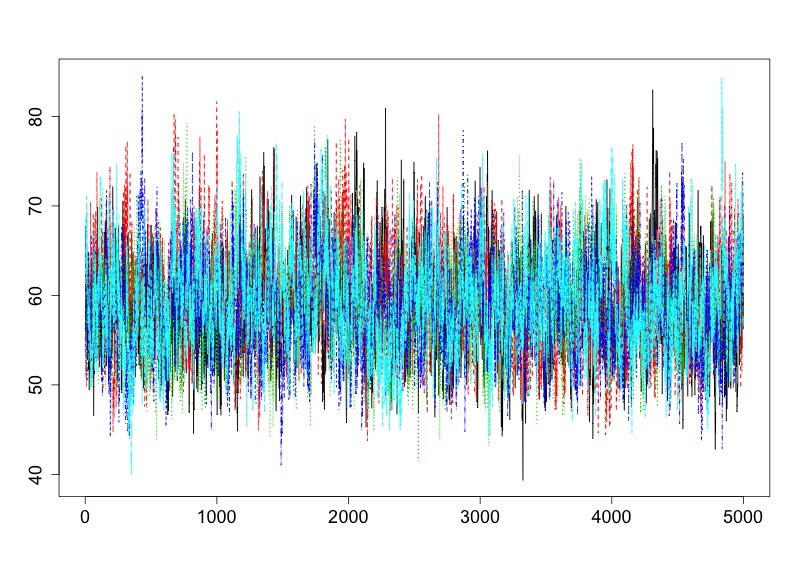}  \\
(a) & (b)\\
\includegraphics[scale=0.28]{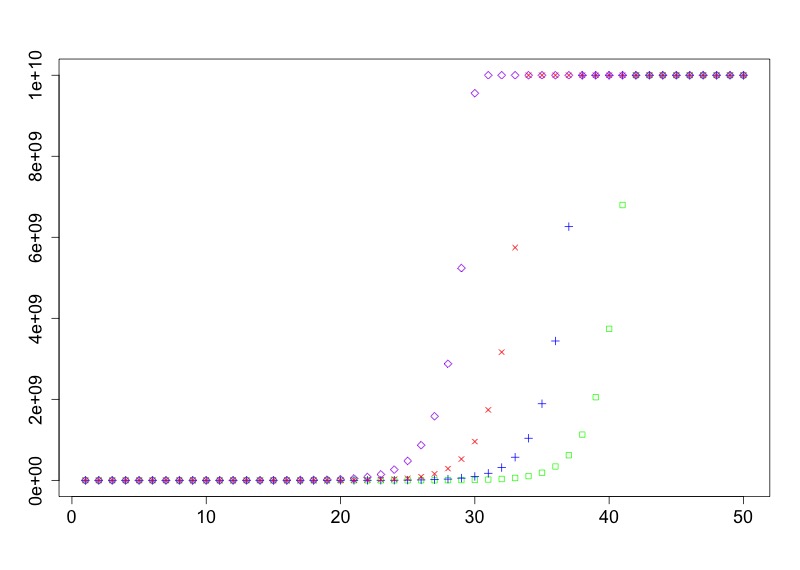} &
\includegraphics[scale=0.28]{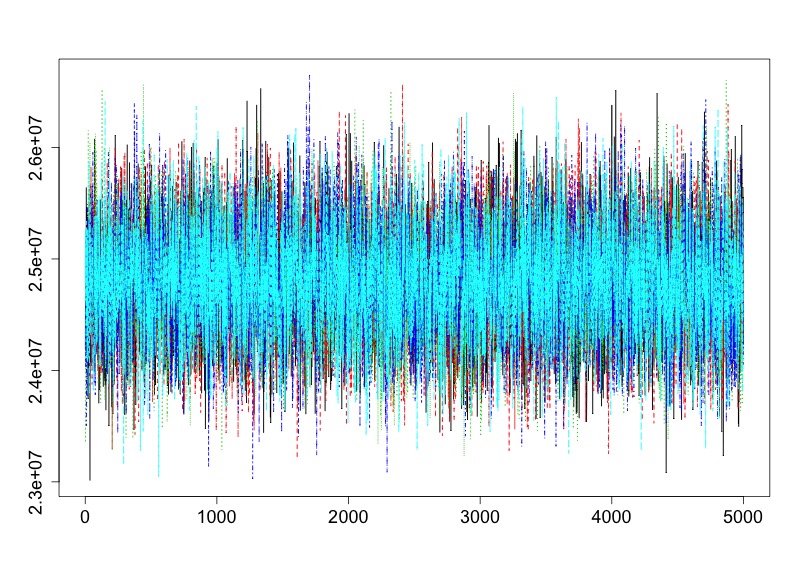} \\
(c) & (d)\\
\end{tabular}
\caption{MCEM output for $\lambda_{1}^{2}$, panel (a) - $d=200$, panel (c) - $d=1500$; Gibbs sampling output for $\lambda_{1}^{2}$, panel (b) - $d=200$, panel (d) - $d=1500$. In each case the horizontal axis represents the iteration number and the vertical axis represents the value of $\lambda_{1}^{2}$ at that iteration.}
\label{comparisons}
\end{figure}

The primary target of inference is the regression coefficients, and these are well-estimated in this case as can be seen from Figure 2, panel (a), where we compare the fully Bayes estimate $E[\vW|\vY]$ to the true values $\vW^{(true)}$. Though not shown, the empirical Bayes estimate $E[\vW|\vY, \hat{\lambda}_{1}^{2}, \hat{\lambda}_{2}^{2}]$ is also accurate in this case.  Moving to the second case where $d=1500$, Figure 1, panel (c), shows the output of the MCEM algorithm $\lambda_{1}^{2}$. In this case the algorithm fails to converge with the value of $\lambda_{1}^{2}$ increasing without bound. Figure 1, panel (d), shows the output of the Gibbs sampler for $\lambda_{1}^{2}$ for five chains initialized at different states, and as with case 1 we see that the Gibbs sampler mixes very well; however, we note that the stationary distribution is shifted towards an \emph{extremely large} value of the tuning parameter, which in a sense corresponds to the behaviour of the MCEM algorithm for this case. The MCEM and Gibbs sampling output for $\lambda_{2}^{2}$ behaves in a manner that is very similar to that of  $\lambda_{1}^{2}$. In terms of the estimation of $\vW$ the result is depicted in Figure 2, panel (b), where we see severe over-shrinkage of the fully Bayes estimate $E[\vW|\vY]$ when compared to the true values, with all of the estimates shrunk very close to zero. The over-shrinkage in this case is a problem directly related to the tuning parameters. To see this, we run the Gibbs sampler with the tuning parameters fixed at their data-generating values, and we compare the resulting estimate $E[\vW|\vY, \lambda_{1_{true}}^{2}=\lambda_{2_{true}}^{2}=2]$ to $\vW^{(true)}$ in Figure 2, panel (c), where the over-shrinkage observed in Figure 2, panel (b), is no longer observed, even though $d=1500$.

\begin{figure*}[htbp]
\centering
\begin{tabular}{cc}
\includegraphics[scale=0.28]{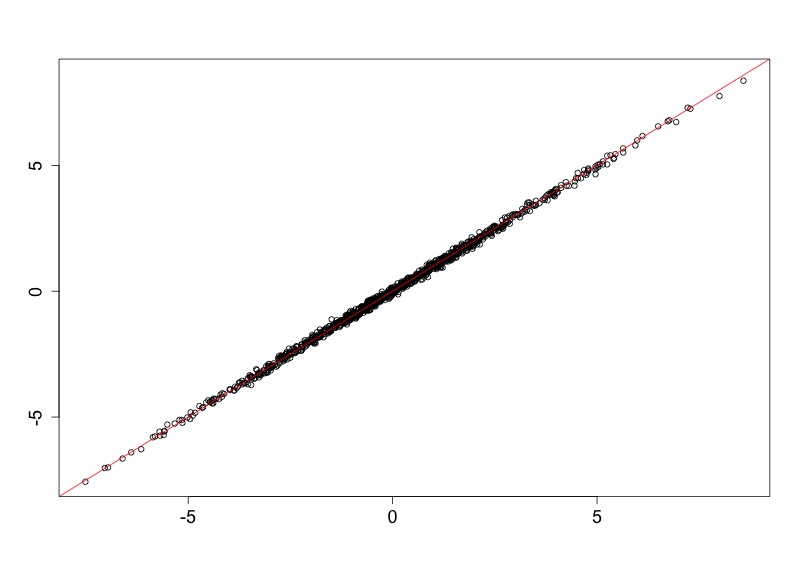}&
\hspace{-1.8em}
\includegraphics[scale=0.28]{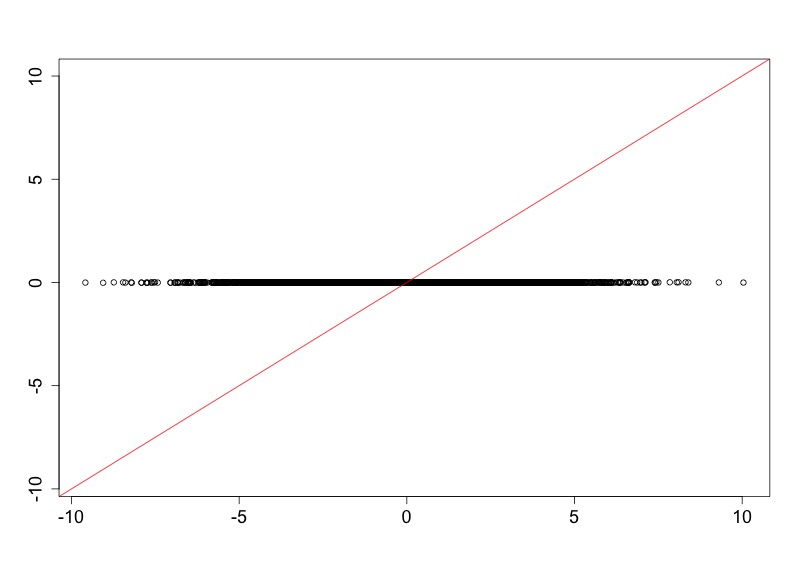}  \\
(a) & (b)\\
\includegraphics[scale=0.28]{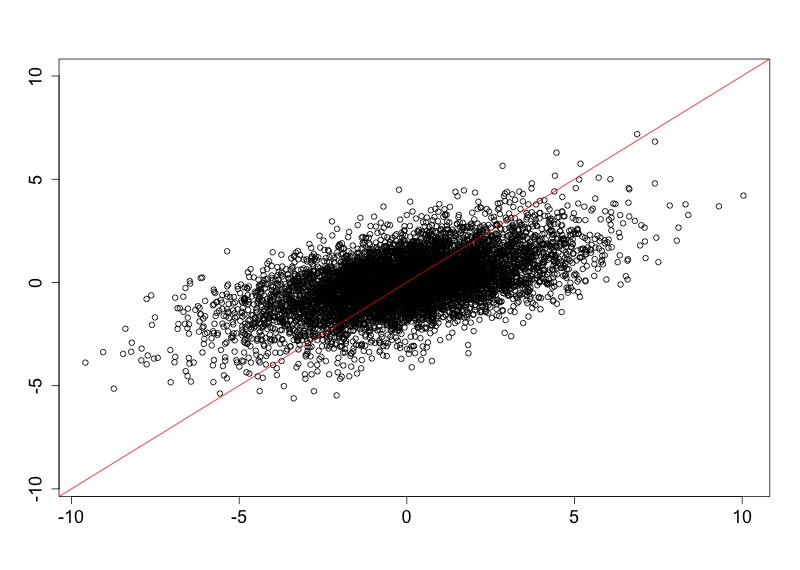} &
\hspace{-1.8em}
\includegraphics[scale=0.28]{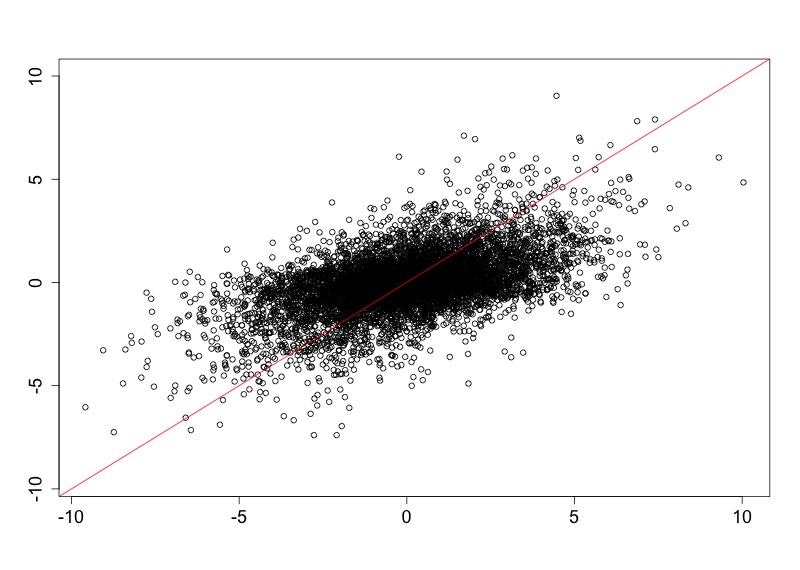} \\
(c) & (d)
\end{tabular}
\caption{Comparison of the posterior mean estimates (vertical axis) $\hat{\vW}$  to the true values (horizontal axis) $\vW$: panel (a) - $d=200$, fully Bayes;  panel (b) - $d=1500$, fully Bayes; panel (c) - $d=1500$, $(\lambda_{1_{true}}^{2},\lambda_{2_{true}}^{2})$; panel (d) - $d=1500$, $(\hat{\lambda}_{1_{WAIC}}^{2},\hat{\lambda}_{2_{WAIC}}^{2})$.}
\label{comparisons}
\end{figure*}

To understand the differences observed in the two cases described above it is instructive to examine the shape of the marginal likelihood $p(\vY|\lambda_{1}^{2},\lambda_{2}^{2}) = \int_\Theta p\: ( \Y , \Theta \; | \lambda_{1}^{2}, \lambda_{2}^{2}) \; d \Theta$ as a function of $\lambda_{1}^{2},\lambda_{2}^{2}$ for each of the two simulated datasets. As with many marginal likelihood calculations this is made difficult by the high-dimensional integral of $p\: ( \Y , \Theta \; | \lambda_{1}^{2}, \lambda_{2}^{2})$ over $\Theta = ( \W , \vtau^{2}, \vomega^{2}, \sigma^{2})$. We therefore develop an approximation to the marginal likelihood and study the shape of this approximation. The approximation to the marginal likelihood is based on first approximating the scale mixing distribution 
(\ref{scale-mix2}) as 
\begin{align}
\begin{split}
p(\vtau^{2},\vomega^{2}|\lambda_{1}^{2},\lambda_{2}^{2}) \approx \prodkK Gamma\left( \tau_k^2 \Big| \left(\frac{m_kc + 1}{2}\right), \left(\frac{\lambda_1^2}{2}\right)\right)  \\\times \prodid Gamma \left( \omega_{i}^{2} \Big| \left(\frac{c + 1}{2}\right), \left(\frac{\lambda_2^2}{2}\right) \right).
\end{split}
\end{align}
This approximation omits the terms of the form $(\tau_{k}^{2}+\omega_{i}^{2})^{-\frac{c}{2}}$ in (\ref{scale-mix2}) and assumes a priori independence between these variables, facilitating the required integration. Using basic properties of the Gaussian distribution the integration over $\W$ can be carried out analytically to yield
$$ \Y | \vtau^{2}, \vomega^{2}, \sig \: \sim \: MVN ( 0, \: (I_c \otimes \mathbf{X}) \vSigma_W (I_c \otimes \mathbf{X}^T ) + \sig I_{cn} \: )$$
where $\vX$ has rows $\vx_{\ell}', \, \ell =1, \dots, n$ and
$$\vSigma_W = \sig I_c \otimes Diag \left\lbrace  \left( \frac{1}{\omei} + \frac{1}{\tau_{k(i)}^2  } \right)^{-1} , \: i =1, \dots, d \right\rbrace. $$ The marginal likelihood can then be expressed as
\begin{align}
\begin{split}
p( \mathbf{Y} | \lamo, \lamt ) =  \int \left[ \int_0^{\infty} p( \Y | \vtau^{2}, \vomega^{2}, \sig) p(\sig) d\sig \right]\\ \times p(\vtau^{2} | \lamo) p(\vomega^{2} | \lamt)   \:   
d \vtau^{2} d \vomega^{2}.   
\end{split}
\end{align}
The inner integral over $\sigma^{2}$ can be carried out analytically after which
the marginal likelihood approximation can be expressed as 
$p(\vY|\lambda_{1}^{2},\lambda_{2}^{2}) = E_{\tau^{2}, \omega^{2}}[p( \Y | \vtau^{2}, \vomega^{2})]$
where the remaining integration over $\vtau^{2}$ and $\vomega^{2}$ is not analytically tractable. We proceed with a simple plug-in approximation\\ $E_{ \; \tau^{2}, \; \omega^{2}} \left[ p( \mathbf{Y} | \vtau^{2} , \vomega^{2}) \right] \; \approx p( \mathbf{Y} | \: E [\vtau^{2}] \: , E[\vomega^{2}] \:)$ where $ E [\tauk] = \frac{m_k c +1}{\lamo}$ and $E[\omei] = \frac{c+1}{\lamt}$ which yields
\begin{small}
\begin{align}
\begin{split}
\label{ML_approx}
p( \mathbf{Y}  | \lamo, \; \lamt) \approx (2 \pi)^{ - \frac{ nc}{2} } b_\sigma^{a_\sigma} \frac{\Gamma( \frac{ nc}{2} + a_\sigma )}{\Gamma( a_\sigma )} \: \times  
\bigg| \vB(\lamt,\lamo) \bigg|^{-\frac{1}{2}}\\
\times \left( b_\sigma + \frac{1}{2} \mathbf{Y}^{T}  \vB(\lamt,\lamo)^{-1} \mathbf{Y} \right)^{ -( \frac{ nc}{2} + a_\sigma ) }, 
\end{split}
\end{align}
\end{small}
where $\vB(\lamt,\lamo) =(I_c \otimes \mathbf{X}) \left(   I_c \otimes \vA(\lamt,\lamo) \right) (I_c \otimes \mathbf{X}^{T}) + I_{cn} \: )$ and 
$$\vA(\lamt,\lamo) = Diag \left\lbrace  \left( \frac{\lamt}{c +1 } + \frac{\lamo}{ m_{k(i)} c + 1} \right)^{-1} \right\rbrace$$.

The marginal likelihood approximation (\ref{ML_approx}) is evaluated over a grid of $(\lambda_{1}^{2},\lambda_{2}^{2})$ for each of the two examples and this is shown in Figure 3, panels (a) and (b), which depict the surface for $d=200$ and $d=1500$, respectively. In panel (a) the approximation appears nicely behaved with a relatively high degree of curvature around the origin and a mode is clearly identified close to the origin. Conversely, the surface depicted in panel (b) is relatively flat across the entire parameter space and a mode is not easily identified. The shape of this surface sheds some light on the behaviour of the MCEM and Gibbs sampling algorithms for this case. Assuming that our approximation is a roughly adequate representation of the shape of the marginal likelihood, it appears that the problems associated with the tuning parameters in certain settings are not computational but rather simply a result of this shape. This problematic behaviour is observed not only when $d>n$ but it can also occur when $d$ is moderately large and the effect sizes are weak. 
\begin{figure}[htbp]
\centering
\begin{tabular}{c}
\includegraphics[scale=0.4]{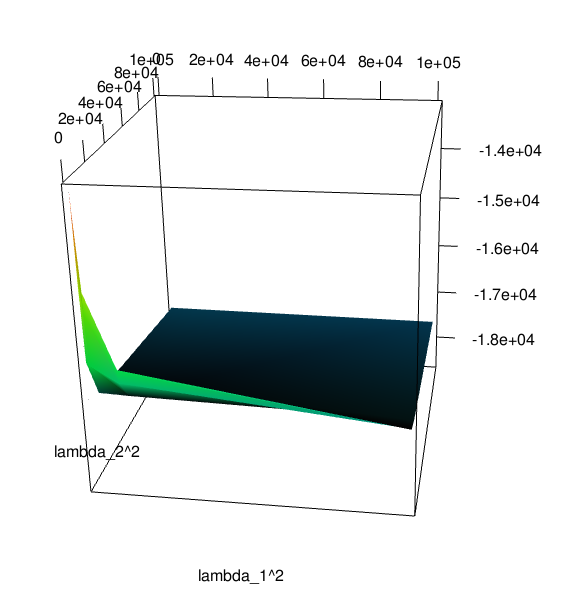} \\
(a)\\
\includegraphics[scale=0.4]{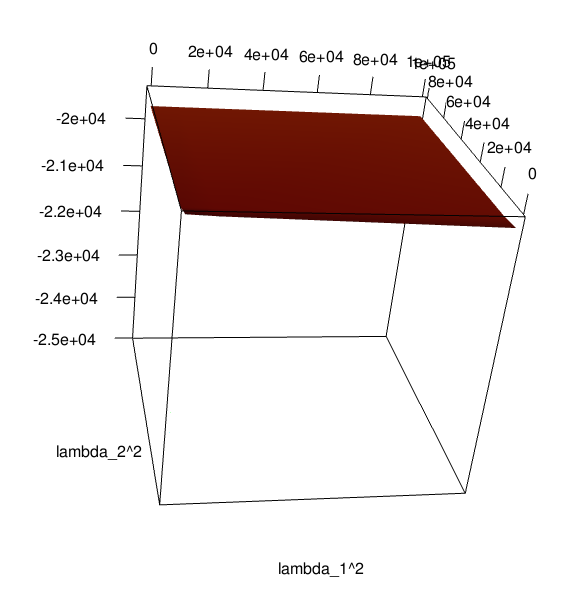} \\
(b) 
\end{tabular}
\caption{Marginal likelihood approximation (log scale) evaluated over a grid of $(\lambda_{1}^{2},\lambda_{2}^{2})$ values: panel (a), $d=200$; panel (b), $d=1500$.}
\label{fig3}
\end{figure}

Interestingly, choosing the tuning parameters using CV or its approximation based on the WAIC does not lead to the same problems we have observed with the fully Bayes and empirical Bayes approaches. This can be seen in Figure 2, panel (d), which compares $E[\vW|\vY, \lambda^{2}_{1_{WAIC}}, \lambda^{2}_{2_{WAIC}}]$ to $\vW^{(true)}$. 

\section{Conclusion}
We have investigated tuning parameter selection for a bi-level Bayesian group lasso model developed for imaging genomics. Through examples we have illustrated problems that arise generally with the fully/empirical Bayes approaches. These problems appear to be caused by the shape of the marginal likelihood. Cross-validation and its approximation based on WAIC do not exhibit the same problems. By imposing \emph{constraints based on out-of-sample prediction}, these approaches avoid the problems that arise from the shape of the marginal likelihood when $d>n$ or when there are only weak effects present. As a simple explanation for this, we note that a solution such as that depicted in Figure 2, panel (b), would not result in good out-of-sample prediction and would therefore not be selected by a CV-based approach. Our recommended approach for choosing the tuning parameters in this Bayesian model is thus to use the WAIC as an approximation to leave-one-out CV. 

\section*{Acknowledgements}
F.S. Nathoo and M. Lesperance are supported by NSERC discovery grants. F.S. Nathoo holds a tier II Canada Research Chair in Biostatistics for Spatial and High-Dimensional Data

\bibliographystyle{plain}

\bibliography{tuning_select}

\end{document}